# I've got the "Answer"!
## Interpretation of LLMs Hidden States in Question Answering

Valeriya Goloviznina[1][0000-0003-1167-2606] and Evgeny Kotelnikov[1][0000-0001-9745-1489]

[1] Vyatka State University, Kirov, Russia
{goloviznínavs, kotelnikov.ev}@gmail.com

**Abstract.** Interpretability and explainability of AI are becoming increasingly important in light of the rapid development of large language models (LLMs). This paper investigates the interpretation of LLMs in the context of the knowledge-based question answering. The main hypothesis of the study is that correct and incorrect model behavior can be distinguished at the level of hidden states. The quantized models LLaMA-2-7B-Chat, Mistral-7B, Vicuna-7B and the MuSeRC question-answering dataset are used to test this hypothesis. The results of the analysis support the proposed hypothesis. We also identify the layers which have a negative effect on the model's behavior. As a prospect of practical application of the hypothesis, we propose to train such "weak" layers additionally in order to improve the quality of the task solution.

## 1    Introduction

Large language models are applied to a wide variety of generative tasks: summarization, machine translation, dialog systems, story generation and code writing [12]. In some tasks, such as the knowledge-based question answering, LLMs already outperform the quality of human answers[1]. However, such models are still not perfect, i.e., not all model's answers are true.

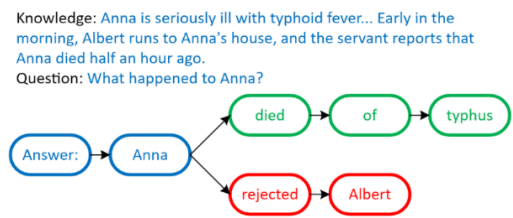

**Fig. 1.** Examples of generation of <span style="color:green">**true**</span> and <span style="color:red">**false**</span> answers to a knowledge-based question.

---

[1]    https://super.gluebenchmark.com/leaderboard, task: MultiRC;
https://russiansuperglue.com/leaderboard/2, task: MuSeRC.

The question arises: at what point does the model make a mistake and deviate from the "correct" behavior necessary to solve the task at hand, which leads to a wrong answer (Fig. 1)? In our work we try to shed light on this problem by interpreting the behavior of the model at the level of hidden states obtained at the output of each of its layers.

The issues of interpretability and explainability have attracted the attention of researchers due to the rapid development of LLMs [15]. Interpretability refers to delving into the decision-making process of the model, increasing the confidence (of developers) in understanding how the model obtains its results. Explainability relates to the ability to provide information (to the user) to build confidence that the AI is making correct and unbiased decisions based on facts [1].

The classification of interpretability and explainability methods is ambiguous. Three review articles on explainable artificial intelligence (XAI) and interpretation of deep neural networks provide three different classifications [1, 8, 10]. In one of these papers, interpretation methods are categorized according to which part of the network they help to interpret: weights, neurons, subnetworks or hidden representations [8].

Our work aims at investigating the interpretability of neural network models, by which we mean revealing its internal properties. We research and interpret models at the level of hidden states.

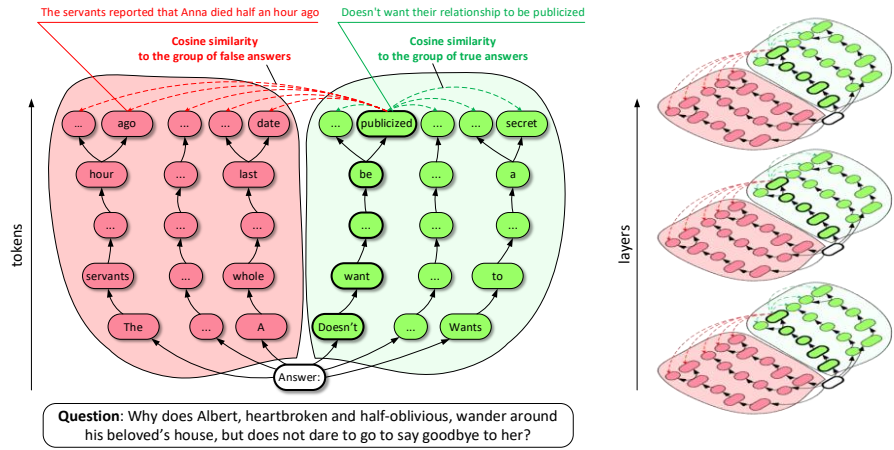

**Fig. 2.** Hypothesis on partitioning the hidden state space of the model. *Left*: the possible LLM's answers to the question are shown in the form of token sequences[2]. Each oval corresponds to a token and its hidden state vector. We consider the hidden state of the last token of some sequence as a vector representation of the entire sequence. We define the similarity between the sequence and the groups of true and false answers as the average cosine similarity between the given sequence and all answers of the group under consideration. *Right*: the described procedure is reproduced on each layer of the model.

---

[2]    For ease of perception, we show tokens as whole words.

Our hypothesis is as follows: correct and incorrect behavior of the model while solving the current task can be distinguished at the level of hidden states. The hidden state space of the model can be divided into two subspaces: hidden states corresponding to correctly generated sequences (in the case of the question answering – correct answers), and hidden states that represent incorrect sequences (wrong answers). Our assumption is reflected in Fig. **2**.

To test the hypothesis, we use quantized versions of the models LLaMA-2-7B-Chat[3], Mistral-7B[4], Vicuna 7B[5], and as a question-answering dataset MuSeRC is used [4]. We define the similarity between a sequence and groups of true and false answers as the average cosine similarity between the current sequence and all answers of the considered group. We confirm the hypothesis by analyzing 200 examples from the MuSeRC dataset. An evolution and practical application of the hypothesis is the suggestion that "weak" layers of the model that have a negative impact on its behavior can be additionally trained in order to improve it.

Our contribution is as follows:

- we propose a hypothesis of partitioning the hidden state space of a model into subspaces corresponding to its correct and incorrect behavior within a certain generative task,
- we propose a procedure for verifying our hypothesis on the basis of analyzing the hidden states of the LLM in a knowledge-based question answering,
- we confirm the hypothesis for three LLMs using the MuSeRC dataset,
- we identify the layers which have a negative effect on the model's behavior.

## 2    Previous work

In this section, we review the works on interpretation of the hidden states of the language models.

Zou et al. [15] proposed a Linear Artificial Tomography (LAT) method to analyze the hidden representations of LLaMA-2-Chat language models. They then control the generation on such aspects as honesty, morality, emotions, harmlessness, memorization and others. The authors show that their approach allows to identify situations where the model lies.

In contrast to [15], the goal of our work is to test the hypothesis of partitioning the hidden state space of the model. In our experiments, in addition to LLaMA-2-7B-Chat, we use the Mistral-7B and Vicuna-7B models.

Yang et al. [13] used hidden states to investigate the influence of input data on model performance. The authors apply the proposed method to analyze the errors of the RoBERTa model [6] in sentiment classification and the occurrence of hallucinations in machine translation of the Transformer model [11].

---

In contrast to [13], we interpret the performance of LLMs on a generative knowledge-based question answering task rather than on classification and translation tasks. The goal of our work is not only to analyze model's errors, but also to determine the specific point at which they occur during the generative process.

Dar et al. [3] projected the hidden states of the model into a set of tokens using logit lens. They described the semantic information flow and revealed patterns in the attention mechanism of the GPT-2 model. To visualize the information flow and the influence of language model components on its output, the authors developed a tool that represents the model as a flow graph, where nodes are neurons or hidden states of the model and edges are interactions between them. They computed the semantic closeness of hidden states projected into a set of tokens to construct information flow.

In our work, we do not map hidden states into the token space, and we compute the similarity between hidden states as cosine similarity vectors not for the purpose of determining changes in semantic flow from layer to layer of the model, but for the purpose of distinguishing correct behavior of the model from incorrect behavior.

Belrose et al. [2] analyzed various autoregressive language models up to 20B parameters in terms of iterative inference, taking into account how the model predictions are refined layer by layer. The tuned lens (improved logit lens) method proposed by the authors also operates on hidden states of the model and their mappings to a set of tokens. This method can be used to detect prompt injection attacks latent in the input data with high accuracy and to identify those parts of the data for which the model requires more training steps.

In contrast to [2], we interpret models using their hidden states directly without additional tools such as logit or tuned lens.

Razova et al. [9] asked whether a language models pay attention to sentiment lexicon when solving the task of text sentiment analysis. For this purpose, the authors studied the attention weight matrices of the Russian-language RuBERT model and conclude that, on average, 3/4 of the attention heads of different variants of the model statistically pay more attention to sentiment lexicon than to neutral lexicon.

In contrast to [9], we interpret the performance of modern LLMs in solving the generative task at the hidden state level.

## 3       Models and dataset

### 3.1       Models

Experiments are conducted with the LLaMA-2-7B-Chat-GPTQ, Mistral-7B-v0.1-AWQ and Vicuna-7B-v1.5-GPTQ quantized models. All models have 32 layers. The choice of models is due to their popularity on the one hand and limited computational resources on the other hand.

Since obtaining and annotating a sufficient variety of true and false answers to knowledge questions is a time-consuming procedure, it was decided to analyze the hidden states of the models based on existing data which is described further.

### 3.2    Dataset

We use the Russian-language MuSeRC dataset [4]. Each example is a text and several questions about that text. Each question has several true and false answers. The questions and answers are written by annotators such that the information of several sentences of the text must be involved to answer the question.

The dataset has 922 examples in total, which contains 5,239 question pairs and answer groups. The answers of 600 examples (training and validation part) with 3,426 questions are labeled into true and false answers. The test part contains 322 examples with answers without labels. The average number of questions per example is 5.7, average number of true answers per question is 1.9, average number of false answers is 2.3. An example of MuSeRC dataset is shown in Appendix A.

We selected the examples that satisfy the following conditions:

1. contain at least 2 true and 2 false answers to each question;
2. the length of each answer is not less than 5 words;
3. the difference between the average lengths of true and false answers does not exceed 30 characters;
4. the answer does not contain a number.

The first condition promotes variety of answers, the second ensures meaningfulness of answers, the third maintains a balance in the length of true and false answers, and the last condition excludes examples in which true and false answers differ by only one number (see Appendix B for an example). As a result, we selected 164 examples containing 217 pairs of questions and answer groups that matched these conditions. The characteristics of examples are given in Table 1.

**Table 1.** Average length (characters) and ROUGE-1 values for texts and answers in the selected examples.

| Avg len of texts | Original answers | | | |
| --- | --- | --- | --- | --- |
| | True | | False | |
| | Avg len | R-1 | Avg len | R-1 |
| 1,294 | 65 | 0.33 | 57 | 0.28 |

For instance, for the example considered in Appendix A, the selected question-answer pair is shown in Table 2.

**Table 2.** Example of selected data from the MuSeRC dataset.

| |
| --- |
| **Text**: text in Appendix A |
| **Question**: Why does Albert, heartbroken and half-oblivious, wander around his beloved's house, but does not dare go to say goodbye to her? |
| **Original true answer**: Doesn't want their relationship to be publicized. |
| **Original true answer**: He is afraid of harming her and himself by publicizing their affair. |
| **Original false answer**: A whole week has passed since their last date. |
| **Original false answer**: The servants reported that Anna died half an hour ago. |

In order to increase the number of true and false answers to 5 for each question, we used the rewriting of answers obtained by GPT-4 Turbo[6]. The prompt format is in Appendix C. For each answer, the model generated 3 rewritten variants. For each variant, the ROUGE-1 value was calculated in relation to each answer of the true or false group. These values were averaged. Rewritten variants were ranked based on average ROUGE-1 scores. Those rewritten variants that increased the diversity of answers, i.e., had the lowest values of this score, were selected to augment the original dataset. We removed 12 examples with a difference between the average length of true and false answers more than 30 characters (the third condition). The characteristics of augmented dataset are given in Table 3.

**Table 3.** Average length (characters) and ROUGE-1 values for texts and answers in the selected examples after rewriting augmentation.

| Avg len of texts | Original answers | | | |
| --- | --- | --- | --- | --- |
| | True | | False | |
| | Avg len | R-1 | Avg len | R-1 |
| 1,294 | 70 | 0.18 | 60 | 0.20 |

The final dataset[7] contains 152 examples, which correspond to 200 pairs of questions and answer groups. Each question-answer pair has 5 true and 5 false answers (Table 4).

**Table 4.** Example of sampled data from the MuSeRC dataset after increasing the number of answers by different rewritten variants.

| |
| --- |
| **Text**: text in Appendix A |
| **Question**: Why does Albert, heartbroken and half-oblivious, wander around his beloved's house, but does not dare go to say goodbye to her? |
| **Original true answer**: Doesn't want their relationship to be publicized. |
| **Original true answer**: He is afraid of harming her and himself by publicizing their affair. |
| **Rewritten true answer**: Prefers to avoid publicity in their relationship. |
| **Rewritten true answer**: Wants to keep their relationship a secret. |
| **Rewritten true answer**: Wants to keep their relationship confidential. |
| **Original false answer**: A whole week has passed since their last date. |
| **Original false answer**: The servants reported that Anna died half an hour ago. |
| **Rewritten false answer**: The service staff reported that Anna died thirty minutes ago. |
| **Rewritten false answer**: The employees informed that Anna's death occurred half an hour ago. |
| **Rewritten false answer**: A message came from the servants that Anna left this world half an hour ago. |

---

[6]    https://platform.openai.com/docs/models/gpt-4-and-gpt-4-turbo

[7]    https://anonymous.4open.science/r/llm_two_subspaces-5CF5

## 4  Experiments

### 4.1  Cosine similarities

To test our hypothesis, we calculated for all the examples the cosine similarity between each answer in the given example and the two groups of answers for that example – correct and incorrect. Answers are represented as token sequences and a hidden state (at given layer) of some sequence is the vector of the last token of this sequence at the considered layer. The similarity between a sequence and a group is the average of the cosine similarity of the given sequence to all sequences in that group.

We obtained hidden states of the models for token sequences containing a prompt with task, knowledge, question, and answer descriptions. The format of the input data is presented in Appendix D.

Thus, three categories of similarities were formed:

- similarity of **true** sequences to one's own group of **true** sequences,
- similarity of **true** (**false**) sequences to another group of **false** (**true**) sequences,
- similarity of **false** sequences to their own group of **false** sequences.

The results of calculating the specified cosine similarity categories at each layer for the example considered in Table 4 are shown in Fig. 3 and Fig. 4 as heatmaps.

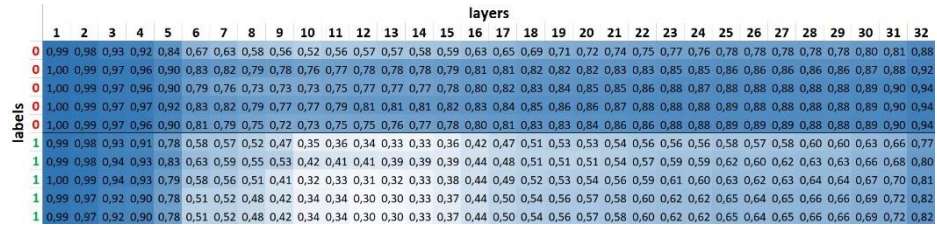

**Fig. 3.** Heatmap of the average cosine similarity values of true and false answers by layer to the **false** answers group for the LLaMA-2-7B-Chat model. White is a low cosine similarity, blue is a high value.

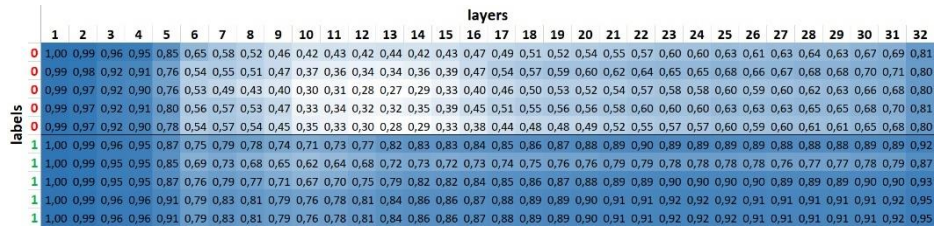

**Fig. 4.** Heatmap of the average cosine similarity values of true and false answers by layer to the **true** answers group for the LLaMA-2-7B-Chat model. White is a low cosine similarity, blue is a high value.

Each row of the table is a true (label=1) or false (label=0) sequence, the columns are the model layers. The value in a cell is the average value of the cosine similarity of the sequence to the group of false (Fig. 3) or true (Fig. 4) sequences.

To analyze the statistics, the obtained cosine similarity values were averaged over the sequences in the group and then averaged over the layers.

For the example from Table 4, these calculations are indicated in Fig. 5. In this example, the average cosine similarity of false sequences to the group of correct sequences is 0.59 (averaged over the sequences in the group and over the layers of the model), and the average cosine similarity of correct sequences to the group of correct sequences is 0.85 (averaged in the same way).

**Fig. 5.** Computation of averaged cosine similarity values over sequences in the group and over model layers for the LLaMA-2-7B-Chat model.

The described procedure for calculating cosine similarities was applied to all 200 question-answer pairs. The average scores for the 200 examples are presented in Table 5.

**Table 5.** Average cosine similarity scores for the three models. Cosine similarity is higher for their own groups and lower for another group.

| The category of cosine similarity | Average cosine similarity | | |
|---|---|---|---|
| | LLaMA-2-7B | Mistral-7B | Vicuna-7B |
| Similarity of true sequences to their own group of true sequences | 0.8240 | 0.8461 | 0.8001 |
| Similarity of true (false) sequences to another group of false (true) sequences | 0.7344 | 0.7666 | 0.7006 |
| Similarity of false sequences to their own group of false sequences | 0.7791 | 0.8076 | 0.7524 |

The distribution of average cosine similarity scores for the 200 examples is shown in Fig. 6.

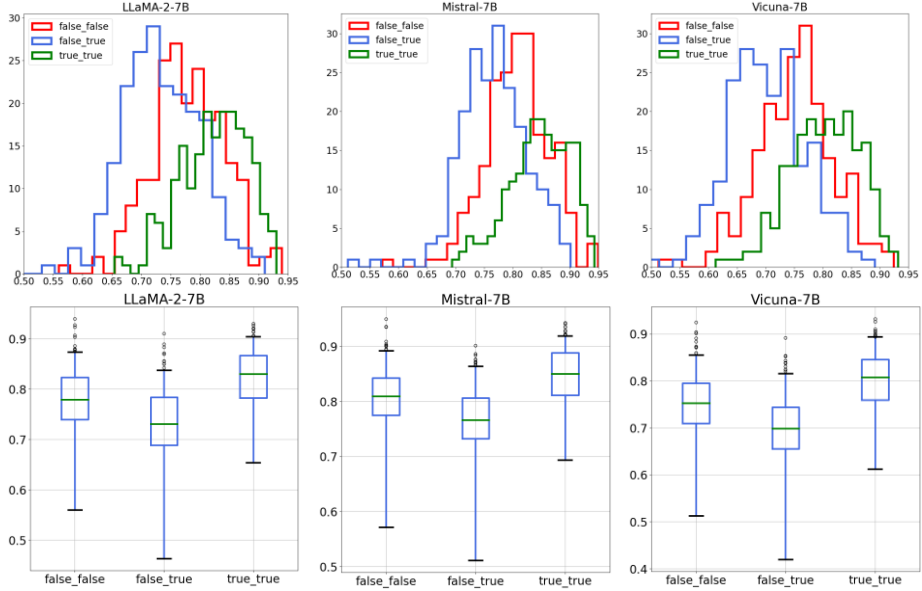

**Fig. 6.** Distribution of average cosine similarity.

## 4.2  Hypotheses testing

To test our hypothesis, two pairs of observations are formed:

- the average values of cosine similarity (averaged over sequences in the group and model layers) of false answers to the groups of false answers and true answers;
- the average values of cosine similarity (averaged in the same way) of true answers to the groups of true and false answers.

Since the observations are independent and their distributions are normal, the t-test can be used for the results of the Mistral-7B and Vicuna-7B models. For the LLaMA-2-7B-Chat model, the Levene test has unequal variances, therefore we used Welch's t-test for it.

We reformulate our hypothesis ("the hidden state space of the model can be divided into two subspaces: correct and incorrect hidden states") as two hypotheses:

- average values of cosine similarity of false answers to the group of false answers are **not** equal to the average values of cosine similarity of false answers to the group of true answers;
- average values of cosine similarity of true answers to the group of true answers are **not** equal to average values of cosine similarity of true answers to the group of false answers.

**Table 6.** t-test

| Model | p-value |
|-------|---------|
| $H_0$: Average values of cosine similarity of false answers to the group of false answers are equal to the average values of cosine similarity of false answers to the group of true answers. | |
| LLaMA-2-7B | $1.69e^{-11}$ |
| Mistral-7B | $1.99e^{-12}$ |
| Vicuna-7B | $3.93e^{-13}$ |
| $H_0$: Average values of cosine similarity of true answers to the group of true answers are equal to average values of cosine similarity of true answers to the group of false answers. | |
| LLaMA-2-7B | $4.77e^{-38}$ |
| Mistral-7B | $5.07e^{-38}$ |
| Vicuna-7B | $6.33e^{-42}$ |

The results of testing these hypotheses are shown in Table 6. All p-values are less than 0.001, thus hypotheses $H_0$ are rejected, that is, there are statistically significant differences between the cosine similarity values of true and false answers to own and other groups.

Thus, we can distinguish in the hidden state space of the model two subspaces – subspace corresponding to correctly generated sequences (true answers), and subspace that represent incorrect sequences (wrong answers).

## 5    Discussion

We analyzed variation of the obtained cosine similarity scores across model layers to identify potentially "weak" layers in need of additional training. Criteria for analyzing layers at the single sequence level are as follows:

- *min_abs*: minimum cosine similarity value out of 32 layers,
- *pos_dif* and *neg_dif*: maximum difference with the previous layer in positive and negative directions.

Criterion for analyzing layers at the sequence group level is as follows:

- *group_dif*: the largest difference between the average similarity of true (false) sequences to another group of false (true) sequences and the average similarity of true (false) sequences to its own group of true (false) sequences.

For the first two criteria, 1,000 (200 examples of 5 true or false sequences each) layer indices were included, and for the third, 200 layer indices were included.

The modes of these layer indices series and their frequencies of occurrence are given in Table 7 and Appendix E.

The distribution of *group_dif* criterion values for the LLaMA-2-7B-Chat model is presented in more detail in the diagrams (Fig. 7, Fig. 8).

**Table 7.** Criteria for analyzing model layers at the sequence group level.

| Model | **False** sequences | | **True** sequences | |
|---|---|---|---|---|
| | mode | freq | mode | freq |
| LLaMA-2-7B | 13 | 32 | 15 | 38 |
| Mistral-7B | 16 | 24 | 16 | 47 |
| Vicuna-7B | 12 | 28 | 15 | 53 |

Given these results, several middle layers of the models require attention and further research. The most frequent minimum value of cosine similarity is found at layer 9 for the Mistral-7B model and at layer 10 for LLaMA-2-7B-Chat and Vicuna-7B. The largest difference between the cosine similarity to the other group relative to one's own group is at layers 12–16.

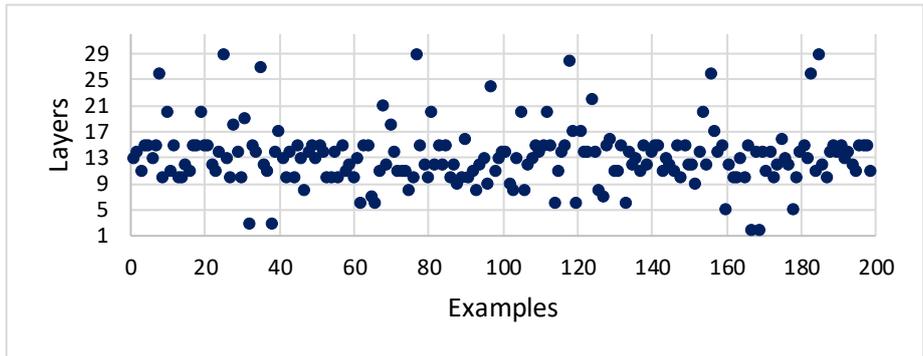

**Fig. 7.** Maximum value by *group_dif* criterion for each example

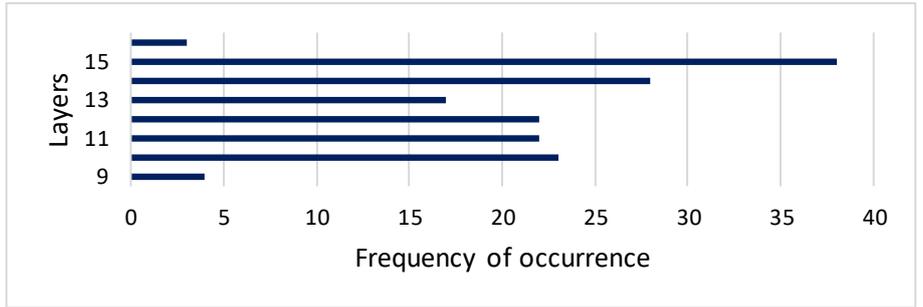

**Fig. 8.** Frequency of occurrence among 200 examples for several middle layer indices (from 9 to 16). By occurrence of the layer index, we mean that the maximum value of *group_dif* is reached at a given layer.

## 6    Conclusion

In this paper we hypothesize that it is possible to distinguish between true and false behavior of the model when solving the current generative task at the level of hidden states, i.e., to divide the space of hidden states of the model into two subspaces: hidden states corresponding to correctly generated sequences (true and false answers in the case of the question-answering task considered in this paper), and hidden states representing incorrect sequences (false answers). The hypothesis is confirmed with three quantized models LLaMA-2-7B-Chat, Mistral-7B and Vicuna-7B on the augmented MuSeRC question-answering dataset. Cosine similarity analysis between groups of correct and incorrect sequences showed that the most likely "weak" layers in the models are the middle layers from 9 to 16.

A development and practical application of the hypothesis is to propose that it is conceivable to additionally train "weak" layers of the model that have a negative effect on its behavior in order to improve it. In addition, the hypothesis should be tested with other different generative tasks for several languages.

## 7    Limitations

In our work, the hypothesis is tested without real sequence generation. This limitation is due to the labor-intensive nature of the annotation process. Despite the focus on generative tasks, our study does not face ethical issues related to these tasks, for the reason mentioned above.

Also, we are limited to analyzing three models that are similar in architecture and dimensionality.

**Acknowledgments.** This work was supported by Russian Science Foundation, project № 23-21-00330, https://rscf.ru/project/23-21-00330/.

**Disclosure of Interests.** The authors have no competing interests to declare that are relevant to the content of this article.

## References

1. Ali, S., Abuhmed, T., El-Sappagh, S., Muhammad, K., Alonso-Moral, J. M., Confalonieri, R., Guidotti, R., Del Ser, J., Díaz-Rodríguez, N., Herrera, F.: Explainable Artificial Intelligence (XAI): What we know and what is left to attain Trustworthy Artificial Intelligence. In Information Fusion, vol. 99 (2023)
2. Belrose, N., Ostrovsky, I., McKinney, L., Furman, Z., Smith, L., Halawi, D., Biderman, S., Steinhardt, J.: Eliciting Latent Predictions from Transformers with the Tuned Lens. arXiv:2303.08112v4, last accessed 2024/04/02
3. Dar, G., Geva, M., Gupta, A., Berant, J.: Analyzing Transformers in Embedding Space. In Proceedings of the 61st Annual Meeting of the Association for Computational Linguistics, vol. 1, pp. 16124–16170. Toronto, Canada (2023)
4. Fenogenova, A., Mikhailov, V., Shevelev, D.: Read and Reason with MuSeRC and RuCoS: Datasets for Machine Reading Comprehension for Russian. In Proceedings of the 28th International Conference on Computational Linguistics, pp. 6481–6497. Barcelona, Spain (2020)

5.  Gunning, D., Stefik, M., Choi, J., Miller, T., Stumpf, S., Yang, G-Z.: XAI–Explainable artificial intelligence. Sci.Robot.4. (2019)
6.  Liu, Y., Ott, M., Goyal, N., Du, J., Joshi, M., Chen, D., Levy, J., Lewis, M., Zettlemoyer, L., Stoyanov, V.: Roberta: A robustly optimized bert pretraining approach. arXiv:1907.11692v1, last accessed 2024/04/02
7.  nostalgebraist, interpreting GPT: the logit lens, https://www.less-wrong.com/posts/AcKRB8wDpdaN6v6ru/interpreting-gpt-the-logit-lens, last accessed 2024/04/02
8.  Räuker, T., Ho, A., Casper, S., Hadfield-Menell, D.: Toward Transparent AI: A Survey on Interpreting the Inner Structures of Deep Neural Networks. In IEEE Conference on Secure and Trustworthy Machine Learning (SaTML), pp. 464–483 (2023)
9.  Razova, E., Vychegzhanin, S., Kotelnikov, E.: Does BERT Look at Sentiment Lexicon? In Recent Trends in Analysis of Images, Social Networks and Texts. In AIST 2021, Communications in Computer and Information Science, vol. 1573, pp. 55–67. Springer, Cham (2022)
10. Tjoa, E., Guan, C.: A survey on explainable artificial intelligence (xai): Toward medical xai. In IEEE Transactions on Neural Networks and Learning Systems, vol. 32, pp. 4793-4813 (2021)
11. Vaswani, A., Shazeer, N., Parmar, N., Uszkoreit, J., Jones, L., Gomez, A. N., Kaiser, Ł., Polosukhin, I.: Attention is all you need. In Advances in Neural Information Processing Systems, NIPS 2017, pp. 6000–6010 (2017)
12. Yang, J., Jin, H., Tang, R., Han, X., Feng, Q., Jiang, H., Yin, B., Hu, X.: Harnessing the power of llms in practice: A survey on chatgpt and beyond. In ACM Trans. Knowl. Discov (2024)
13. Yang, S., Huang, S., Zou, W., Zhang, J., Dai, X., Chen, J.: Local Interpretation of Transformer Based on Linear Decomposition. In Proceedings of the 61st Annual Meeting of the Association for Computational Linguistics, vol. 1, pp. 10270–10287. Toronto, Canada (2023)
14. Zhao, H., Chen, H., Yang, F., Liu, N., Deng, H., Cai, H., Wang, S., Yin, D., Du, M.: Explainability for Large Language Models: A Survey. In ACM Transactions on Intelligent Systems and Technology (2024)
15. Zou, A., Phan, L., Chen, S., Campbell, J., Guo, P., Ren, R., Pan, A., Yin, X., Mazeika, M., Dombrowski, A.-K., Goel, S., Li, N., Byun, M. J., Wang, Z., Mallen, A., Basart, S., Koyejo, S., Song, D., Fredrikson, M., Kolter, J. Z., Hendrycks, D.: Representation Engineering: A Top-Down Approach to AI Transparency. arXiv:2310.01405v3, last accessed 2024/04/02

## Appendix A

An example[8] from MuSeRC dataset that contains 6 question-answer pairs:

| **idx**: 481 |
| --- |
| **text[9]**: Albert goes to Anna's house and sees that all the lights are turned off and only a ray of light breaks through her window. How can I find out what's wrong with her? A saving thought occurs to him that in the event of her illness, he can check on her health through a messenger, and the messenger does not necessarily need to know who gave him the order. So he learns that Anna is seriously ill with typhoid fever and her illness is very dangerous. Albert suffers unbearably at the thought that Anna could be dying now, and he cannot see her before her |

---

[8]  This example and examples below are translated from Russian into English.

[9]  In the original dataset, the sentences in the text are numbered. The numbering is removed using regular expressions.

death. But he does not dare to rush upstairs to his beloved even now, for fear of harming her and himself by publicizing their affair. Heartbroken and half-oblivious, Albert wanders around his beloved's house, not daring to go to her to say goodbye. A week has passed since their last date. Early in the morning, Albert runs to Anna's house, and the servants report that Anna died half an hour ago. Now the painful hours of waiting for Anna seem to him the happiest of his life. And again the hero lacks the courage to enter the rooms, and he returns an hour later, hoping to blend in with the crowd and remain unnoticed. On the stairs he encounters unfamiliar mourning people, and they only thank him for his visit and attention.

| question | original answers | | | |
|---|---|---|---|---|
| | **true** | | **false** | |
| What happened to Anna? | Anna died of Typhoid. | Anna suffered from Typhoid and died of fever. | Anna ran away with Albert. | Anna rejected Albert. |
| Why didn't Albert go to say goodbye to Anna? | He was afraid to show that they were having an affair. | He was afraid of harming her reputation. | He missed the train. | He was hampered by unforeseen circumstances. |
| What illness did Albert learn about Anna? | Typhoid fever. | Typhoid fever. | Cold. | Flu. |
| How many minutes after Anna's death did Albert come? | Half an hour. | Thirty. | Eight. | Ten. |
| How does Albert find out that Anna is seriously ill with typhoid fever and her illness is very dangerous? | Through a messenger. | He inquired about her health through a messenger. | Came to her. | From passersby. |
| Why does Albert, heartbroken and half-oblivious, wander around his beloved's house, but does not dare go to say goodbye to her? | Doesn't want their relationship to be publicized. | He is afraid of harming her and himself by publicizing their affair. | A whole week has passed since their last date. | The servants reported that Anna died half an hour ago. |

## Appendix B

An example of an instance of the MuSeRC dataset excluded from further work due to a difference between answers of only one number.

```
    "id": 397,
    "text": "(1) The Norwegian men's national biathlon team won the relay race … (13) They
were ahead of their main rivals - the Germans - by only 0.3 seconds.",
    "questions": [{
        "question": "How many seconds were the women's team ahead of their rivals?",
        "answers": [
            {"text": "By 0.3 seconds.",
             "label": 1
            },
```

```
        …
            {"text": "By 0.5 seconds.",
             "label": 0
           }],
        "idx": 0
    }]
```

## Appendix C

GPT-4 rewriting prompt.

Paraphrase the text. Write 3 different rewritten variants. To write the answer, use the following structure:

Rewriting:
#1# Variant 1
#2# Variant 2
#3# Variant 3
Text: {answer}
Rewriting:

## Appendix D

Input data format during getting the hidden states of the model.

[INST] <<SYS>>
You can answer questions in Russian, based on the data provided.
<</SYS>>
Briefly answer the question using the knowledge given to you. [/INST]
Knowledge: {knowledge}
Question: {question}
Answer:

## Appendix E

Criteria for analyzing model layers at the level of single sequences.

| Criterion | min_abs | | pos_dif | | neg_dif | |
|---|---|---|---|---|---|---|
| | mode | freq | mode | freq | mode | freq |
| **LLaMA-2-7B** | | | | | | |
| | | | group of **true** sequences | | | |
| **false** sequences | 10 | 371 | 31 | 619 | 5 | 900 |
| | | | group of **false** sequences | | | |
| **true** sequences | 10 | 400 | 31 | 669 | 5 | 929 |
| **Mistral-7B** | | | | | | |
| | | | group of **true** sequences | | | |
| **false** sequences | 9 | 412 | 31 | 529 | 3 | 493 |
| | | | group of **false** sequences | | | |
| **true** sequences | 9 | 484 | 31 | 561 | 1 | 488 |
| **Vicuna-7B** | | | | | | |
| | | | group of **true** sequences | | | |
| **false** sequences | 10 | 193 | 31 | 824 | 4 | 584 |
| | | | group of **false** sequences | | | |
| **true** sequences | 10 | 218 | 31 | 857 | 4 | 565 |